\documentclass{article}
\usepackage{ijcai26}

\usepackage{times}
\usepackage{soul}
\usepackage{url}
\usepackage[hidelinks]{hyperref}
\usepackage[utf8]{inputenc}
\usepackage[small]{caption}
\usepackage{graphicx}
\usepackage{amsmath}
\usepackage{amsthm}
\usepackage{booktabs}
\usepackage{algorithm}
\usepackage{algpseudocode}
\usepackage{siunitx}
\usepackage{amsmath,amssymb,amsfonts,amsthm, mathtools}
\usepackage{mathtools}
\usepackage{xspace}
\usepackage{bbm}
\usepackage{caption}
\usepackage{subcaption}
\usepackage{enumitem}
\usepackage{xcolor}

\usepackage{tikz}
\usetikzlibrary{positioning}

\newcommand{\mm}{{SAVGO}}

\usepackage{latexsym}

\title{SAVGO: Learning State–Action Value Geometry with Cosine Similarity for Continuous Control}

\author{
Stavros Orfanoudakis
\and
Pedro P. Vergara\\
\affiliations 
Delft University of Technology, The Netherlands\\
\emails
\{s.orfanoudakis, p.p.vergarabarrios\}@tudelft.nl
}

\begin{document}

\maketitle

\begin{abstract}
While representation and similarity learning have improved the sample efficiency of Reinforcement Learning (RL), they are rarely used to shape policy updates directly in the action space.
To bridge this gap, a geometry-aware RL algorithm that explicitly incorporates value-based similarity into the policy update, State–Action Value Geometry Optimization (SAVGO), is proposed. In detail, SAVGO learns a joint state–action embedding space in which pairs with similar action-value estimates exhibit high cosine similarity, while dissimilar pairs are mapped to distinct directions. This learned geometry enables the generation of a similarity kernel over candidate actions sampled at each update, allowing policy improvement to be guided directly toward higher-value regions beyond local gradient-based updates. As a result, representation learning, value estimation, and policy optimization are unified within a single geometry-consistent objective, while preserving the scalability of off-policy actor–critic training. The proposed method is evaluated on standard MuJoCo continuous-control benchmarks, demonstrating improvements over strong baselines on challenging high-dimensional tasks. Ablation studies are done to analyze the contributions of value-geometry learning and similarity-based policy updates.
\end{abstract}

\section{Introduction}
\label{sec:intro}

Deep Reinforcement Learning (RL) has demonstrated remarkable success across a wide range of domains, from Atari games to continuous-control robotics~\cite{mnih2015humanlevel,schulman2017ppo}.
However, in realistic high-dimensional environments, training is still frequently limited by sample inefficiency and instability.
Off-policy actor-critic methods, such as Soft Actor Critic (SAC) and Twin Delayed Deep Deterministic Policy Gradient (TD3), are strong RL algorithms for continuous control due to their scalability and robustness \cite{haarnoja2018sac,fujimoto2018td3}. More recent algorithms further improve sample efficiency through critic ensembles and distributional objectives \cite{kuznetsov2020tqc,chen2021redq}.
Even with these improvements, performance often remains highly sensitive to representation quality and hyperparameter choices, while function approximation can obscure or distort the local decision structure needed for reliable policy improvement.

A prominent line of work addresses these limitations by improving representations learned from observations. In pixel-based control, data augmentation and contrastive objectives have substantially boosted data efficiency \cite{yarats2021drq,laskin2020curl,chen2020simclr}, while self-predictive objectives leverage temporal structure to regularize latent dynamics \cite{schwarzer2021spr}.
Latent-dynamics auxiliary objectives have also been studied through reward and transition prediction in compressed state spaces \cite{gelada2019deepmdp}.
More recently, return and value aware auxiliary tasks aim to align features with decision-making signals rather than purely visual invariance \cite{liu2021rcrl,yue2023vcr}.
State--action representation learning has also been used to improve actor--critic learning in both online and offline continuous-control settings~\cite{fujimoto2023sale}, suggesting that explicitly modeling interactions between states and actions can provide useful structure for value estimation and policy learning. Related work has also explored value-consistent and reward-aware representation objectives that more directly couple learned features to downstream control performance \cite{yue2023vcr,rewardaware2025proto}.
Despite these advances, most methods still treat representation learning as an auxiliary component, where the policy improvement step typically remains a local gradient update that does not explicitly exploit the geometry of the learned latent space.

\begin{figure}[t]
\includegraphics[width=0.8\linewidth]{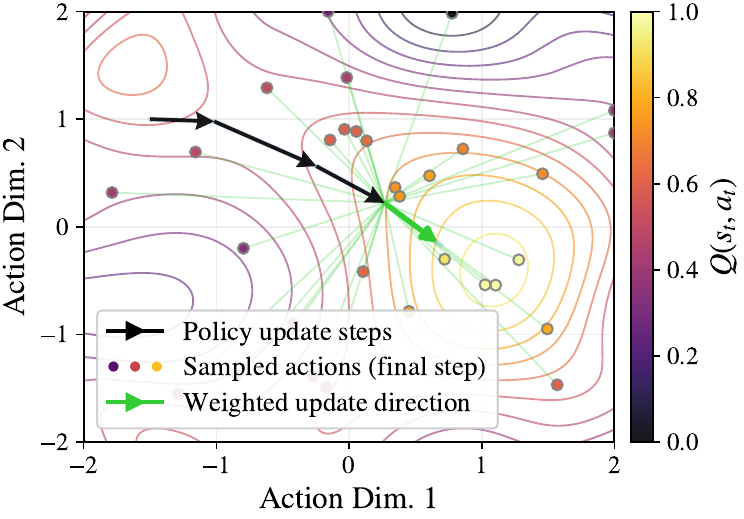}
\centering
\caption{Policy improvement with value-aware geometry for a single state. Colored contours depict the critic landscape $Q(s_t,\cdot)$. Candidate actions (colored dots) are sampled and reweighted using the learned cosine-similarity geometry, resulting in a kernel-aggregated update direction (green) and stable policy update steps (black) that lead to higher-value regions beyond the pointwise gradient.}
\label{fig:dist_space overview}
\end{figure}

Distance and similarity-based approaches offer a complementary perspective by defining behavioral notions of similarity and learning representations that preserve them. Bisimulation-inspired objectives relate latent distances to control-relevant equivalence classes, yielding more robust representations \cite{zhang2021dbc,kemertas2021robustbisim}. Sampling-based state-similarity objectives further improve scalability by enabling direct optimization from experience \cite{castro2021mico}.
More recent work has extended this direction through action-based and controllability-oriented representations, predictive or variational bisimulation metrics, and large-scale analyses of behavioral metric learning under distractors \cite{liang2026taskaware,freed2026variationalbisim}.
In parallel, episodic memory and nearest-neighbor methods exploit similarity to accelerate value estimation and improve learning dynamics \cite{pritzel2017nec,shen2021nnac}.
This idea has been extended in continuous control by using neighboring state--action structure, nonparametric value estimates, or action-value discrepancy information to improve policy exploitation \cite{gao2026ira}.
Despite these advances, a key limitation remains, as distances are primarily used to regularize representations or shape value targets and only rarely serve as first-class operators for policy improvement in the action space. As a result, policy updates remain largely local, gradient-driven, and weakly informed by the global structure of the learned value function, contributing to instability and inefficient exploration in high-dimensional action spaces.

To address this limitation, we propose \textit{State–Action Value Geometry Optimization} (\mm), a geometry-aware RL algorithm that explicitly integrates value-based similarity into the policy update itself. \mm~ learns a joint state–action embedding space in which pairs with similar state-action $Q$-values exhibit high cosine similarity, while dissimilar pairs are mapped to different directions. In parallel, this learned geometry can be used to create a similarity kernel over $K$ candidate actions sampled at each update, enabling the policy to be directly pulled toward regions associated with higher values (Figure~\ref{fig:dist_space overview}). In this way, \mm~tightly couples representation learning, value estimation, and policy improvement through a single geometry-consistent objective, while retaining the scalability and practicality of replay-based off-policy actor–critic training. 
We evaluate \mm~on standard continuous-control benchmarks from MuJoCo, comparing against strong RL baselines and demonstrating consistent gains on challenging, high-dimensional tasks, such as Humanoid. In addition, we conduct targeted ablation studies to isolate the contributions of the value-geometry objective, similarity-based kernel weighting, and key hyperparameters. The main contributions of this work can be summarized as:
\begin{itemize}
  \item Proposing \mm\, an RL algorithm that learns a cosine-similarity geometry over state-action embeddings to distinguish between similar and dissimilar pairs based on their utility difference, rather than using representation learning only as an auxiliary regularizer.
  \item Deriving a geometry-aware policy optimization operator that performs similarity-weighted aggregation over candidate actions.
  \item Empirically evaluating \mm\ on continuous-control benchmarks, including comparisons to established RL baselines and targeted ablations isolating the effects of the value-geometry objective, kernel weighting, and key temperature/curvature hyperparameters.
\end{itemize}

\section{Related Work}
\label{sec:related_prelim}

\subsection{Representation Learning for RL}
Learning task-relevant representations has long been recognized as a key ingredient for stable and sample-efficient deep RL. For visual control, contrastive objectives combined with augmentation have been particularly effective. SimCLR popularized simple instance discrimination in vision \cite{chen2020simclr}, and CURL adapted contrastive learning to off-policy RL by training an encoder jointly with the control objective \cite{laskin2020curl}. Related augmentation-driven regularization methods, such as DrQ, further improved data efficiency from pixels \cite{yarats2021drq}. Beyond contrastive learning, predictive self-supervision has also been employed to stabilize representation learning, e.g., by predicting future latent features with target encoders \cite{schwarzer2021spr}. Several works have advocated for decoupling representation learning from policy optimization using strong unsupervised objectives \cite{stooke2021decoupling}.

A central limitation of many such objectives is that they are often generic and do not explicitly encode control-relevant structure. This has motivated value-aware representation learning methods, including the VIP perspective \cite{dabney2021vip}, return-based contrastive learning \cite{liu2021rcrl}, and value-consistent representation learning \cite{yue2023vcr}. 
Representation learning in low-dimensional continuous-control domains has also been used, where observations are already compact, but the interaction between states and actions can still benefit from learned structure. In particular, SALE learns state--action embeddings for deep RL and integrates them into TD3 to form TD7, demonstrating that state--action representation learning can substantially improve both online and offline continuous-control performance \cite{fujimoto2023sale}. This line of work is closely related to SAVGO because it highlights the importance of modeling state--action interactions rather than learning state-only features.
SAVGO follows the same overarching goal of making representations control-relevant, but differs in how the learned structure is used. Rather than treating representation learning as an auxiliary regularizer or using it only to improve value targets, a cosine-similarity geometry over \emph{state--action} embeddings is learned and then reused directly to construct similarity-weighted policy updates.

\subsection{Distances, Metrics, and State Abstraction}

Distances over states (and state--action pairs) provide a principled route to representation learning and state abstraction, as behavioral similarity can be formalized by requiring similar rewards and similar future evolution. This idea is captured by bisimulation metrics, in which states are considered close when their immediate rewards and transition dynamics match under an appropriate coupling. DeepMDP connects latent dynamics modeling objectives to bisimulation-style guarantees by learning representations that support reward and transition prediction in latent space \cite{gelada2019deepmdp}. Deep Bisimulation for Control (DBC) minimizes a bisimulation metric loss to learn invariant representations and improve robustness to distractors \cite{zhang2021dbc}. MICo introduces a practical sampling-based behavioral distance that scales to deep RL and uses it to shape representations \cite{castro2021mico}.
Furthermore, action-bisimulation encoding can learn controllability-oriented representations through a recursive invariance constraint, allowing representations to capture long-horizon action-relevant structure without relying directly on reward supervision~\cite{rudolph2024actionbisim}. Predictive and variational bisimulation objectives have also been explored as mechanisms for learning task-relevant or distractor-robust latent spaces \cite{liang2026taskaware,freed2026variationalbisim}. In parallel, large-scale empirical studies of behavioral metric learning have examined how well metric-based objectives filter distractors and have highlighted the importance of design choices in bridging theoretical guarantees and practical deep RL performance \cite{luo2025behavioralmetric}.

Despite their appeal, bisimulation-driven objectives can be brittle in practice, exhibiting robustness issues and representation pathologies such as collapse or exploding norms \cite{kemertas2021robustbisim}. Additional pitfalls have been reported in offline settings, where missing transitions can bias metric estimation and degrade learned distances \cite{zang2023pitfalls}. In contrast to these methods, SAVGO does not aim to learn a full behavioral metric for state abstraction, instead, a value-aware cosine geometry over \emph{state--action} embeddings is learned.

\subsection{Kernel and Nearest-Neighbor Value Search}

A separate but closely related field uses similarity directly in value estimation. Kernel-Based Reinforcement Learning approximates Bellman backups using kernel regression in continuous state spaces \cite{ormoneit2002kbrl}. Neural Episodic Control stores a memory of past embeddings with fast-updated value estimates, effectively implementing a differentiable nearest-neighbor value function \cite{pritzel2017nec}. Nearest Neighbor Actor-Critic proposes a theoretically grounded nearest-neighbor module that can replace or augment parametric value networks \cite{shen2021nnac}, whereas
Instant Retrospect Action (IRA), uses Q-representation discrepancy evolution to learn discriminative representations for neighboring state--action pairs, together with $k$-nearest-neighbor action-value estimates \cite{gao2026ira}.
Collectively, these methods demonstrate that learned similarity can support effective bootstrapping and credit assignment.
Building on this insight, \mm~extends the role of similarity beyond value estimation to policy optimization. Rather than using distances solely to interpolate or retrieve value estimates, \mm~learns a control-relevant state–action geometry and employs it directly as the driving signal for policy improvement.


\section{State–Action Value Geometry Optimization}
\label{sec:method}

Building on the limitations of existing representation-learning, similarity-based, and kernel-driven approaches discussed above, \mm~is introduced as a geometry-aware off-policy actor–critic framework in which learned value-based similarity is made directly actionable during policy improvement. \mm~is composed of two tightly coupled components: 
(i) a \emph{state–action value geometry} learned in a normalized embedding space using cosine similarity, and
(ii) a \emph{geometry-aware policy improvement operator} that aggregates a set of candidate actions through similarity-dependent weights.
In the remainder of this section, the value-geometry learning objective and the resulting similarity-based policy improvement operator are formally derived.

\subsection{Preliminaries: Off-Policy Actor--Critic}
\label{sec:prelim}

A discounted Markov decision process (MDP) $\mathcal{M}=(\mathcal{S},\mathcal{A},P,r,\gamma)$ is considered,
where $\mathcal{S}$ and $\mathcal{A}$ denote the state and action spaces, $P(s_{t+1}\!\mid s_t,a_t)$ denotes the
transition probabilities, $r:\mathcal{S}\!\times\!\mathcal{A}\!\to\!\mathbb{R}$ is the reward function, and $\gamma\in[0,1)$ is the discount factor. A stochastic policy $\pi_\theta(a\mid s)$, parameterized by $\theta$ is used, and transitions are stored in a replay buffer $\mathcal{D}$.
The training loop is aligned with SAC \cite{haarnoja2018sac}, in which, two critics $Q_{\phi_1},Q_{\phi_2}$ are trained using slowly soft-updated target networks $Q_{\tilde\phi_1},Q_{\tilde\phi_2}$.
Following common practice, the minimum over the two target critics is used to obtain conservative value estimates~\cite{fujimoto2018td3}, i.e., $Q(s_t,a_t) \triangleq \min_{m\in{1,2}} Q_{\tilde{\phi}m}(s_{i},a_{i})$.

Accordingly, the critics are updated as in SAC. 
Given a minibatch $\{(s_i,a_i,r_i,s_{i+1},d_i)\}_{i=1}^B \sim \mathcal{D}$ of size $B$, where $d_i$ refers to a terminal state flag, a next action $a_{i+1}\sim\pi_\theta(\cdot\mid s_{i+1})$ is sampled and the temporal difference (TD) target is:
\begin{equation}
\label{eq:sac_target}
y_i
=
r_i+\gamma \; d_i \Big(Q(s_{i+1},a_{i+1})
-\eta \log \pi_\theta(a_{i+1}\mid s_{i+1})
\Big)
\end{equation}
where $\eta\ge 0$ denotes the entropy temperature~\cite{haarnoja2018sac}.
Each critic is then updated by minimizing
\begin{equation}
\label{eq:critic_loss}
\mathcal{L}_{Q_m}(\phi_m)
=
\frac{1}{B}\sum_{i=1}^B
\Big(Q_{\phi_m}(s_i,a_i)-y_i\Big)^2,
\; m\in\{1,2\}.
\end{equation}
All remaining SAC components (reparameterized sampling, temperature tuning, and target updates) are adopted from \cite{haarnoja2018sac} and are not restated. 

\subsection{Value-Aware State--Action Geometry Learning}
\label{sec:geometry}

\begin{figure}[t]
  \centering
  \includegraphics[width=0.9\linewidth]{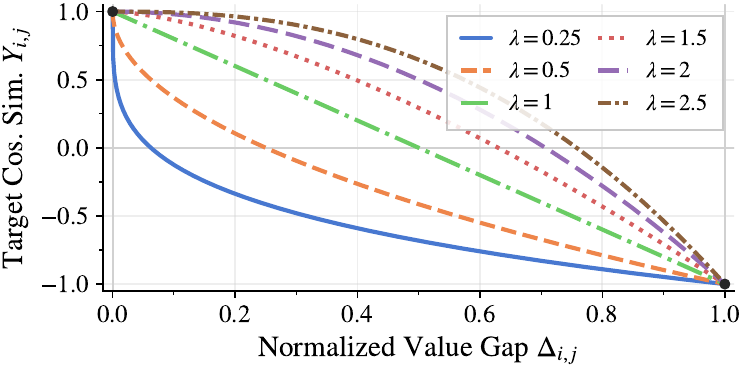}
\caption{Target cosine similarity $Y_{i,j}$ as a function of the normalized value gap $\Delta_{i,j}$ for different curvature parameters $\lambda$.}
  \label{fig:rep-geometry}
\end{figure}

A value-aware state--action geometry is learned to capture control-relevant relationships between pairs of state--action tuples $(s_i,a_i)$ and $(s_j,a_j)$. To this end, a representation encoder function $z_\psi:\mathcal{S}\times\mathcal{A}\rightarrow\mathbb{R}^d$,
parameterized by $\psi$, is trained such that similarity in the embedding space reflects similarity in expected utility. For notational convenience, we denote $\zeta_i \triangleq z_\psi(s_i,a_i)$ and $\zeta_j \triangleq z_\psi(s_j,a_j)$. Similarity between embeddings is measured using cosine similarity,
\begin{equation}
\label{eq:cos_sim}
\cos\!\left(
\zeta_i,
\zeta_j
\right)
=
\frac{
\zeta_i^\top \zeta_j
}{
\|\zeta_i\|_2 \, \|\zeta_j\|_2
} \;\;\;\;\in\;
[-1,1].
\end{equation}
Cosine similarity is adopted due to its bounded range, scale invariance, and compatibility with normalized embeddings, which together yield a stable and interpretable similarity measure on the unit hypersphere.

To ensure that embedding similarities reflect value proximity between state--action pairs, the representation function $z_\psi$ is trained using value-based supervision. For a pair of state--action tuples with action-value estimates $Q(s_i,a_i)$ and $Q(s_j,a_j)$, a normalized value gap $\Delta_{i,j}\in[0,1]$ is defined as
\begin{equation}
\label{eq:gap_norm}
\Delta_{i,j}
=
\mathrm{clip}\!\left(
\frac{|Q(s_i,a_i) - Q(s_j,a_j)|}{\beta},\,0,\,1
\right),
\end{equation}
where $\beta>0$ is a robust scale parameter, such as an exponential moving average of observed value differences. This normalization ensures invariance to reward scaling and prevents large value discrepancies from dominating the learned value geometry.

With $\Delta_{i,j}$ then providing a normalized measure of value dissimilarity on $[0,1]$, it can be directly mapped to a \emph{target cosine similarity} so that the supervision signal matches the range and semantics of cosine similarity, i.e., $+1$ for highly similar pairs and $-1$ for strongly dissimilar pairs. This is achieved with the bounded curvature transformation
\begin{equation}
\label{eq:target_sim}
Y_{i,j}
=
1 - 2\,(\Delta_{i,j})^{\lambda}
\;\;\;\;\;\in\;
[-1,1],
\end{equation}
with $\lambda>0$ controlling the curvature of this mapping and thus the resolution assigned to different gap regimes. As illustrated in Figure~\ref{fig:rep-geometry}, smaller values of $\lambda$ allocate more dynamic range to small gaps, sharpening discrimination among near-optimal actions, while larger values compress small gaps and yield smoother geometries by reducing sensitivity to minor value differences.

Finally, the encoder parameters $\psi$ are optimized by matching predicted similarities $\cos(\zeta_i,\zeta_j)$ to their targets $Y_{i,j}$, using common regression losses $\ell(\cdot)$, such as the Huber or $\ell_2$ loss.
\begin{equation}
\label{eq:rep_loss}
\mathcal{L}_{z}(\psi)
\triangleq
\ell\!\left(
\cos(\zeta_i,\zeta_j) - Y_{i,j}
\right) 
\;\;\;\;\forall (i,j) \in \mathcal{B}.
\end{equation}
Through this construction, the learned embedding space becomes explicitly control-relevant, because state--action pairs with similar utility are encouraged to occupy nearby directions on the unit hypersphere, while pairs with dissimilar utility are pushed apart. This value-aware geometry forms the foundation for the similarity-based policy improvement operator introduced in the following section.

\subsection{Policy Improvement in Learned Geometry}
\label{sec:kernel_pi}

\begin{figure}[t]
\includegraphics[width=1\linewidth]{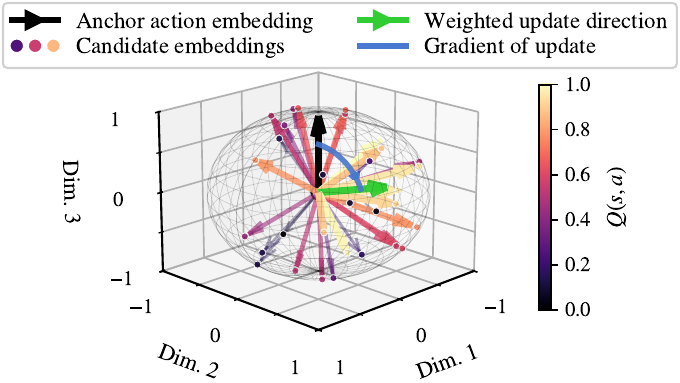}
\centering
\caption{Unit-sphere view of the geometry-aware actor update in the embedding space for state $s_t$.
The anchor embedding $z(s_t,\hat a_t)$ (black) and candidate embeddings $z(s_t,a_t^k)$ (colored by $Q(s_t,a_t^k)$) lie on the learned embedding space.
Cosine similarities to the anchor define kernel weights $w_k$, yielding a similarity-weighted update direction (green) that steers the actor toward high Q-value actions.
}
\label{fig:overview}
\end{figure}

Policy improvement is performed by leveraging the value-aware geometry from Section~\ref{sec:geometry} to construct similarity-weighted updates in the action space. An abstract view is provided in Figure~\ref{fig:dist_space overview}, while Figure~\ref{fig:overview} illustrates the same mechanism on the learned unit-sphere embedding space.

In detail, for a given state $s_t$, an \emph{anchor action} $\hat{a}_t \sim \pi_\theta(\cdot \mid s_t)$ is first sampled from the current policy and mapped to its embedding $\hat{\zeta}_t \triangleq z_\psi(s_t,\hat{a}_t)$ (black vector in Figure~\ref{fig:overview}). In addition, a set of $K$ candidate actions $\{a_t^k\}_{k=1}^K$ is drawn from a proposal distribution (e.g., the policy or a mixture for broader coverage), yielding candidate embeddings $\zeta_t^k \triangleq z_\psi(s_t,a_t^k)$ (colored vectors). Each candidate pair $(s_t,a_t^k)$ is scored by a conservative critic estimate, $q_t^k = \min_{m\in{1,2}} Q_{\tilde{\phi}_m}(s_t,a_t^k)$, which is showed by the vector coloring in Figure~\ref{fig:overview}.

The influence of each candidate on the update is determined by its \emph{geometric proximity} to the anchor in the learned cosine space. Cosine similarities $ \cos(\hat{\zeta}_t,\zeta_t^k) \triangleq \cos(z_\psi(s_t,\hat{a}_t), z_\psi(s_t,a_t^k))$ quantify how close a candidate lies to the anchor direction on the unit sphere. As illustrated in Figure~\ref{fig:overview}, these similarities are later converted into kernel weights, so that nearby candidates contribute more strongly to the aggregated update direction than distant ones. For additional stability, similarity computations can be performed with a slowly-updated target encoder $z_{\tilde\psi}$.

The pairwise similarities are then transformed into a probability simplex through a temperature-controlled softmax,
yielding a similarity kernel over candidate actions,
\begin{equation}
\label{eq:kernel_weights}
w_k
=
(1-\varepsilon)\,
\frac{\exp(cos(\hat{\zeta}_t,\zeta_t^k)/\rho)}{\sum_{j=1}^K \exp(cos(\hat{\zeta}_t,\zeta_t^j)/\rho)}
+
\frac{\varepsilon}{K},
\end{equation}
where the temperature $\rho>0$ controls the sharpness of the kernel and $\varepsilon \in [0,1]$ prevents weight collapse. In our implementation, $\rho$ is adapted over training using a cosine schedule from $\rho_{\max}$ to $\rho_{\min}$, producing smoother weighting early on while allowing sharper selection later when the representation encoder is more informative.

Using the weights $\{w_k\}_{k=1}^K$, a similarity-weighted value estimate is computed as,
\begin{equation}
\label{eq:q_tilde}
\hat{Q}(s_t,\hat a_t)
\triangleq
\sum_{k=1}^K w_k \, q_t^k
\end{equation}
This can be viewed as a local, similarity-weighted average of critic values around $\hat a_t$ in the learned geometry (Figure~\ref{fig:dist_space overview} and \ref{fig:overview}).
This aggregation smooths the update signal by combining multiple candidates rather than relying on a single point estimate, compared to classic algorithms such as SAC and TD3.
Finally, the policy parameters $\theta$ are updated by minimizing
\begin{equation}
\label{eq:actor_loss}
\mathcal{L}_{\pi}(\theta)
\triangleq
\mathbb{E}_{s_t,\hat a_t \sim \pi_\theta(\cdot \mid s_t)}
\Big[
\eta \log \pi_\theta(\hat a_t \mid s_t)
-
\hat Q(s_t,\hat a_t)
\Big],
\end{equation}
where $\eta$ controls the strength of entropy regularization, thereby promoting exploration in maximum-entropy stochastic actor--critic methods, such as SAC.

From this perspective, policy improvement is no longer driven by a pointwise gradient at a single action,
but by a similarity-weighted aggregation over a set of candidate actions.
This approach is related to weighted policy improvement and advantage-weighted regression methods shown in \cite{peters2010reps,abdolmaleki2018mpo}, but differs fundamentally in that the weights are determined by
learned geometric similarity in state--action space, rather than solely by action-value magnitude.



\begin{algorithm}[t]
\caption{State-Action Value Geometry Optimization}
\label{alg:savgo}
\begin{algorithmic}[1]
\State Initialize actor $\pi_\theta$, critics $Q_{\phi_m}$, encoder $z_\psi$ 
\State Initialize target net. $\tilde\psi\leftarrow\psi$ and $\tilde\phi_m\leftarrow\phi_{m}$, \; {\small $m\in\{1,2\}$}
 
\For{$t=1,2,\dots$}
    \State Sample $a_t\!\sim\!\pi_\theta(\cdot|s_t)$
    \State Step env and  store $(s_t,a_t,r_t,s_{t+1},d_t)\!\in\!\mathcal{D}$
    \State Sample minibatch $\{(s_i,a_i,r_i,s_{i+1},d_i)\}_{i=1}^B \sim \mathcal{D}$

    \State{\textbf{Critics Update}}
    \State  \;\; Compute targets $y_i$ (Eq.~\ref{eq:sac_target})
    \State  \;\; Update $\phi_m$ by $\nabla_{\phi_m}\mathcal{L}_{Q_m}$ (Eq.~\ref{eq:critic_loss}) \;\; for $m\in\{1,2\}$

    \State{\textbf{Value-geometry Encoder Update}}
    \State  \;\; Form pairs $(i,j)$ from the minibatch
    \State  \;\; Compute $\Delta_{ij}$ (Eq.~\ref{eq:gap_norm}) and $Y_{ij}$ (Eq.~\ref{eq:target_sim})
    \State  \;\; Update $\psi$ by $\nabla_\psi \mathcal{L}_z$ (Eq.~\ref{eq:rep_loss})

    \State {\textbf{Actor Update} (for each $s_i$)}
    \State  \;\; Sample anchor $\hat a_i\!\sim\!\pi_\theta(\cdot|s_i)$
    \State  \;\; Sample candidates $\{a_i^k\}_{k=1}^K\!\sim\!\mu(\cdot|s_i)$
    \State  \;\; Compute kernel weights $w_{k}$ (Eq.~\ref{eq:kernel_weights}) using $z_{\tilde\psi}$
    \State  \;\; Compute $\widehat Q(s_i,\hat a_i)$ (Eq.~\ref{eq:q_tilde})
    \State  \;\; Update $\theta$ by $\nabla_\theta \mathcal{L}_\pi$ (Eq.~\ref{eq:actor_loss})

    \State{\textbf{Target Networks Update}}
    \State  \;\; $\tilde\phi_m \leftarrow \tau\,\tilde\phi_m + (1-\tau)\,\phi_m$ \;\; for $m\in\{1,2\}$
    \State  \;\; $\tilde\psi \leftarrow \tau\,\tilde\psi + (1-\tau)\,\psi$
\EndFor
\end{algorithmic}
\end{algorithm}

\subsection{Overall Algorithm and Training Procedure}
\label{sec:algorithm}

Having introduced the value-aware geometry objective (Section~\ref{sec:geometry}) and the geometry-aware policy improvement operator (Section~\ref{sec:kernel_pi}), the complete training procedure is summarized in Algorithm~\ref{alg:savgo}.

The method follows a standard replay-based off-policy actor--critic loop, while coupling critic learning.
At each environment interaction step, an action is sampled from the current stochastic policy and executed (lines~4--5), and the resulting transition is stored in the replay buffer (line~5). A minibatch is then drawn from the buffer (line~6) and used for the parameter updates. The critics are first trained with a SAC-style TD target (lines~7--9), yielding conservative estimates via the minimum of two critics. The value-geometry encoder is updated using paired samples from the same minibatch (lines~10--13) by regressing cosine similarities toward bounded targets derived from value gaps, so that nearby embedding directions reflect similar expected utility. The geometry-aware policy improvement is performed (lines~14--19) by sampling, for each state $s_i$, an anchor action $\hat a_i\sim\pi_\theta(\cdot \mid s_i)$ (line~15) and $K$ candidate actions (line~16), converting anchor--candidate cosine similarities into a soft kernel (line~17), forming a kernel-aggregated value estimate $\widehat Q(s_i,\hat a_i)$ (line~18), and updating the actor to increase this similarity-weighted value signal (line~19) rather than relying on a single pointwise estimate.
Finally, target networks are updated via Polyak averaging (soft-updates) with rate $\tau$ (lines~20--22) to stabilize value and similarity estimations.



\section{Experimental Evaluation}
\label{sec:experiments}

This section evaluates \mm~ on continuous-control benchmarks, comparing against strong baselines and analyzing the effect of key design choices through targeted ablations.

\subsection{Experimental Setup}
\label{sec:exp_setup}

Experiments are conducted on the standard MuJoCo~\cite{towers2025gymnasiumstandardinterfacereinforcement} continuous-control suite (v5) under a fixed budget of $1$M environment steps per run, spanning tasks of varying action dimensionality and difficulty. \mm\footnote{The implementation code is available at~\url{https://github.com/StavrosOrf/DistanceRL}.}~is compared against widely used continuous-control baselines, including Proximal Policy Optimization (PPO)~\cite{schulman2017ppo}, TD3~\cite{fujimoto2018td3}, SAC~\cite{haarnoja2018sac}, and TQC~\cite{kuznetsov2020tqc}, which collectively represent strong on-policy and off-policy training pipelines and are sufficient to characterize performance across the suite. Distance-based representation methods such as MICo~\cite{castro2021mico} and DBC~\cite{zhang2021dbc} are also closely related. However, in our adaptation of the original codebases to the present training stack, MICo and DBC methods did not achieve competitive performance on the more complex MuJoCo tasks and are therefore not included as primary baselines.

\begin{figure*}[t]
  \centering  
  \includegraphics[width=0.95\linewidth]{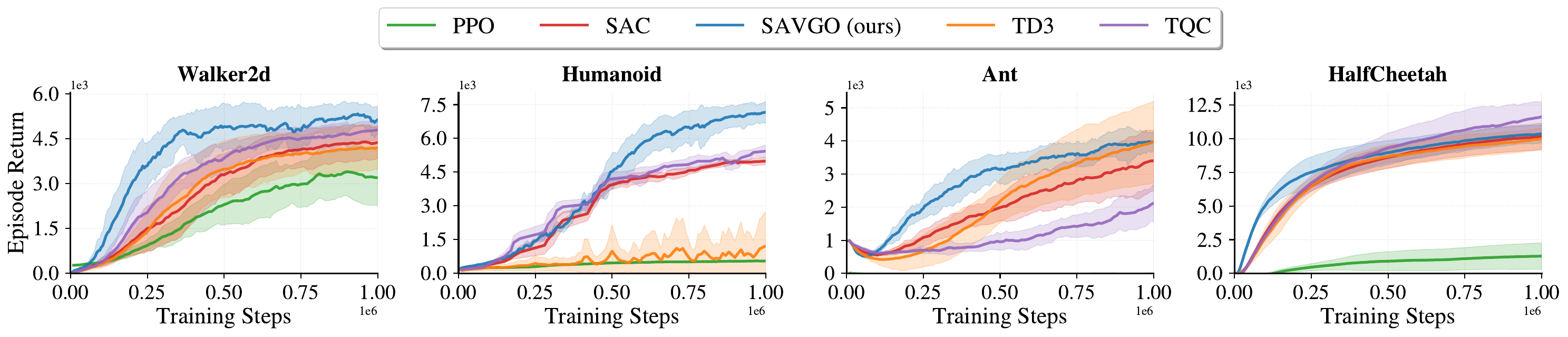}
    \caption{Training curves over $1$M environment steps on representative MuJoCo (v5) tasks. Solid lines show the mean evaluation return over seeds, and shaded regions indicate the standard deviation. The baseline implementations from Stable Baselines 3 are used.}    

  \label{fig:results_training}
\end{figure*}

\begin{table*}[t]
\centering
\small
\caption{Best evaluation performance (maximum over training up to 1{,}000{,}000 steps) across MuJoCo v5 environments. Entries report mean $\pm$ standard deviation over 10 seeds. Bold indicates the best-performing algorithm per environment. The \textit{Total Rewards} row sums environment-wise means, its $\pm$ value is the sum of standard deviations across environments (for a rough aggregate variability indicator).}

\label{tab:results}
\begin{tabular}{l
                S[table-format=-5.0] @{\,$\pm$\,} S[table-format=4.0]
                S[table-format=-5.0] @{\,$\pm$\,} S[table-format=4.0]
                S[table-format=-5.0] @{\,$\pm$\,} S[table-format=4.0]
                S[table-format=-5.0] @{\,$\pm$\,} S[table-format=4.0]
                S[table-format=-5.0] @{\,$\pm$\,} S[table-format=4.0]}
\toprule
\textbf{Environment (v5)} &
\multicolumn{2}{c}{\textbf{PPO}} &
\multicolumn{2}{c}{\textbf{TD3}} &
\multicolumn{2}{c}{\textbf{SAC}} &
\multicolumn{2}{c}{\textbf{TQC}} &
\multicolumn{2}{c}{\textbf{SAVGO}} \\
\midrule

Ant
& 12 & 72
& 4366 & 1296
& 4086 & 789
& 3041 & 668
& \bfseries 4653 & \bfseries 235 \\

HalfCheetah
& 2363 & 1022
& 10259 & 838
& 10388 & 1069
& \bfseries 12114 & \bfseries 1006
& 10545 & 616 \\

Hopper
& 3527 & 284
& \bfseries 3584 & \bfseries 77
& 3371 & 84
& 3571 & 65
& 3573 & 114 \\

Humanoid
& 554 & 62
& 799 & 266
& 5351 & 75
& 6330 & 227
& \bfseries 7687 & \bfseries 319 \\

Inverted2DPendulum
& 9350 & 16
& 9358 & 2
& \bfseries 9360 & \bfseries 0
& \bfseries 9360 & \bfseries 0
& \bfseries 9360 & \bfseries 0 \\

InvertedPendulum
& \bfseries 1000 & \bfseries 0
& \bfseries 1000 & \bfseries 0
& \bfseries 1000 & \bfseries 0
& \bfseries 1000 & \bfseries 0
& \bfseries 1000 & \bfseries 0 \\

Reacher
& -3 & 0
& -3 & 0
& \bfseries -2 & \bfseries 0
& \bfseries -2 & \bfseries 0
& -3 & 0 \\

Swimmer
& \bfseries 114 & \bfseries 25
& 102 & 46
& 77 & 10
& 106 & 25
& 88 & 28 \\

Walker2d
& 4321 & 635
& 4503 & 752
& 4693 & 572
& 4952 & 374
& \bfseries 5626 & \bfseries 747 \\

\midrule
\textbf{Total Rewards}
& 21238 & 2116
& 33968 & 3277
& 38324 & 2599
& 40472 & 2365
& \bfseries 42529 & \bfseries 2059 \\
\bottomrule
\end{tabular}
\end{table*}

To ensure a fair comparison, all methods are trained under identical environment settings and step budgets and are evaluated with the same protocol. Implementations are based on \textit{Stable Baselines 3}~\cite{sb3}. Results are aggregated over multiple random seeds, and performance is reported as mean and standard deviation. Observation normalization is enabled across tasks to reduce sensitivity to scale and improve training stability. Two-layer actor, critic, and encoder networks with 256 nodes per layer are used across algorithms. For \mm, the in-state candidate set size is set to $K\in\{64,\dots,256\}$ depending on action dimension, with kernel smoothing $\varepsilon=0.05$.
SAVGO's encoder network consists of two layers, each with 256 nodes.
The weight smoothing temperature $\rho$ is cosine annealed over the first 200,000 training steps with $\rho_\text{max}=0.75$ and $\rho_\text{min}=0.05$. 
Detailed hyperparameter settings for each environment and baseline can be found in the supplementary material.
All the experiments were run on the ``Anonymous'' supercomputer, consisting of Nvidia V100 GPUs and Intel Xeon 48-core CPU nodes.


\subsection{Main Results on MuJoCo}
\label{sec:main_results}

Figure~\ref{fig:results_training} reports evaluation episode return over $1$M environment training steps averaged over 10 seeds. Across environments, faster early learning and smoother convergence were observed for \mm~relative to PPO, TD3, SAC, and TQC, indicating improved sample-efficiency and a more stable policy-improvement signal. The clearest gains appeared in the higher-dimensional tasks. On \textit{Walker2d}, \mm~learned rapidly and stabilized at a higher return band than the baselines, while on \textit{Humanoid}, it was the only method to consistently achieve strong improvement over training, whereas the others progressed more slowly and/or plateaued earlier. Competitive behavior was also observed on \textit{Ant} and \textit{HalfCheetah}, where \mm~tracked the top-performing baselines while exhibiting reduced variance across training.
Overall runtimes stayed practical even on the hardest MuJoCo tasks (1M training steps): PPO $<2$h, TD3 $\approx2$h, SAC $\approx2.5$h, TQC $\approx4.5$h, and SAVGO $\approx8$h. SAVGO was the slowest due to extra critic and encoder evaluations for similarity weighting, while PPO was the fastest, followed by TD3 and SAC.

Across MuJoCo benchmarks, the best evaluation returns within the 1M-step budget (Table~\ref{tab:results}) show that \mm~is consistently among the strongest methods and achieves the best overall performance, with the largest gains concentrated on challenging high-dimensional tasks such as \textit{Humanoid} and \textit{Walker2d}, where policy improvement is particularly sensitive to critic noise. On mid-complexity environments (e.g., \textit{Ant}), improvements remain clear but smaller, while on \textit{HalfCheetah} performance is competitive rather than best. As expected, on simpler tasks where near-ceiling behavior is reached, differences largely vanish. Overall, these trends suggest that similarity-weighted updates are most beneficial when action ranking is difficult and local gradients are less reliable.


\subsection{Ablation Studies}
\label{sec:ablations}

Ablation studies can quantify the contribution and sensitivity of key \mm~design choices, with a focus on the representation curvature parameter $\lambda$ (Eq.~\ref{eq:target_sim}) and the number of candidate actions $K$ used in the kernelized update (Eq.~\ref{eq:kernel_weights}). Evaluations were performed on \textit{HalfCheetah} and \textit{Humanoid}, which were selected to contrast a lower-dimensional, fast-learning locomotion task, and a high-dimensional setting with higher critic noise and action-ranking difficulty.
Each configuration was trained for $1$M environment steps over 5 random seeds, and performance was summarized by the maximum episode return achieved within the budget.

\begin{figure}[t]
\centering
  \includegraphics[width=\linewidth]{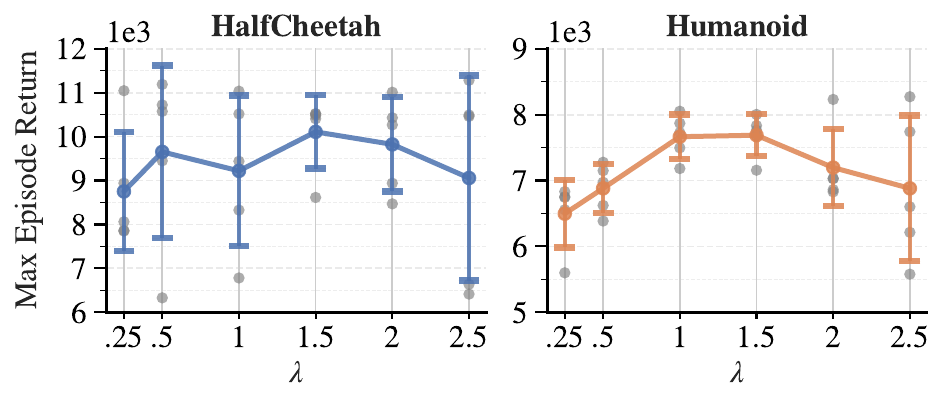}
\caption{Sensitivity to the representation curvature parameter $\lambda$. Points denote per-seed maxima and error bars show mean$\pm$std of the maximum evaluation return within 1M steps over 5 seeds.}

\label{fig:gamma_sweep1}
\end{figure}

\paragraph{Effect of representation curvature $\lambda$.}
A focused ablation isolates the effect of the representation curvature parameter $\lambda$ in Eq.~\ref{eq:target_sim}, while keeping the architecture, optimization settings, replay configuration, and all other \mm~hyperparameters fixed. Figure~\ref{fig:gamma_sweep1} shows a strong dependence on curvature, with values close to 1 yielding the most reliable performance. Small curvature values ($\lambda\le0.5$) increase seed variance, consistent with an overly sharp target mapping that over-weights small value gaps and destabilizes geometry learning (See Figure \ref{fig:rep-geometry}). Large curvature values ($\lambda\ge2$) reduce peak value gaps, suggesting that the mapping becomes too smooth to preserve informative action-ranking structure. Overall, $\lambda\approx 1$--$1.5$ offers the best trade-off between stability and final performance across both tasks.

\begin{figure}[t]
\centering
\includegraphics[width=\linewidth]{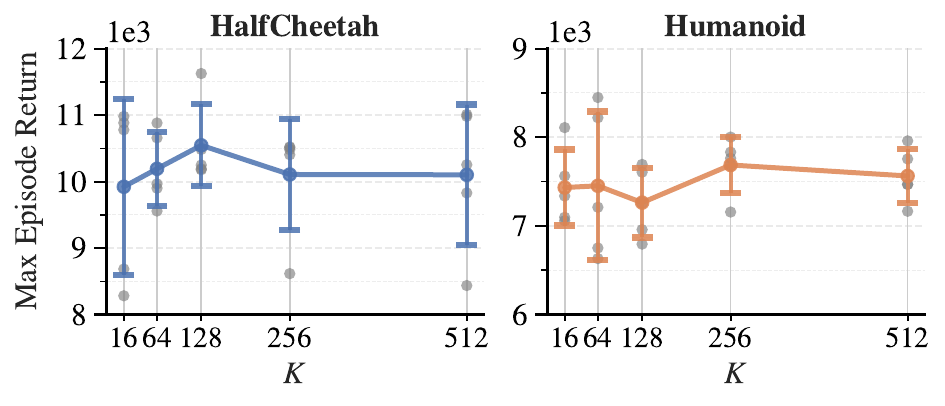}
\caption{Sensitivity to candidate set size $K$ in the similarity-weighted policy improvement operator. Points denote per-seed maxima and error bars show mean$\pm$std of the maximum evaluation return within 1M environment steps over 5 seeds.}

\label{fig:gamma_sweep2}
\end{figure}

\paragraph{Effect of sample candidates $K$.}
\label{sec:ablate_K_mu}
The ablation here studies the effect of the candidate set size $K$ in the kernelized policy improvement operator (Eq.~\ref{eq:kernel_weights}) by sweeping $K\in\{16,64,128,256,512\}$ while holding all other hyperparameters fixed. Figure~\ref{fig:gamma_sweep2} indicates that strong performance is typically obtained with a moderate number of candidates. On \textit{HalfCheetah}, returns remain largely unchanged for $K\ge 64$, since adding more samples doesn't benefit anymore once the local neighborhood is sufficiently covered. On \textit{Humanoid}, smaller candidate sets ($K=16$) reduce peak performance and increase variability, whereas intermediate values ($K\approx 128$) yield the most consistent results. Increasing to $K=512$ does not produce systematic gains, suggesting that additional candidates become redundant and mainly increase computation. Overall, $K$ is best chosen to balance neighborhood coverage and efficiency, with larger action spaces typically requiring larger candidate sets.

\paragraph{Design choice ablation.}
To quantify the contribution of each design choice in \mm~ and verify that the intended policy-improvement mechanism is realized in practice, a targeted ablation study is performed.
Kernel sharpness is ablated by disabling the adaptive temperature schedule in Eq.~\ref{eq:kernel_weights} and keeping $\rho$ fixed. Value-aware geometry learning is ablated by freezing the encoder $z_\psi$, i.e., stopping gradients from Eq.~\ref{eq:rep_loss}, which tests whether similarity must be learned to track value proximity rather than used as a fixed heuristic. Robust normalization is ablated by removing the adaptive scaling in Eq.~\ref{eq:gap_norm} and using a fixed $\beta$, which tests sensitivity to critic's value magnitude. Finally, a stress-test replaces the similarity kernel with uniform weights $w_k=1/K$ to verify that gains come from geometry-aware weighting rather than unweighted averaging.
As anticipated, Table~\ref{tab:ablation} shows that each component was necessary for strong performance on \textit{HalfCheetah} and \textit{Humanoid} after $1$M steps (5 seeds). Fixing $\rho$ or removing adaptive $\beta$ scaling reduced returns, and freezing the encoder (removing Eq.~\ref{eq:rep_loss}) caused a larger drop, indicating that the value-aware geometry must be learned rather than treated as a fixed heuristic. The stress test collapsed performance (uniform weights), confirming that policy improvement strictly depends on similarity-based weighting.

\begin{table}[t]
\centering
\small
\caption{Design choice ablation study on two MuJoCo environments.
Each row removes or modifies one architectural component of the proposed method.
Results are reported as mean evaluation return $\pm$ standard deviation over 5 seeds after 1M steps.}
\label{tab:ablation}
\begin{tabular}{l
                S[table-format=5.0] @{\,$\pm$\,} S[table-format=4.0]
                S[table-format=4.0] @{\,$\pm$\,} S[table-format=3.0]}
\toprule
\textbf{Variant} &
\multicolumn{2}{c}{\textbf{HalfCheetah}} &
\multicolumn{2}{c}{\textbf{Humanoid}} \\
\midrule

SAVGO (full)
& \bfseries 10546 & \bfseries 616
& \bfseries 7687  & \bfseries 319 \\

w/o adaptive $\rho$
& 9184 & 941
& 7559 & 667 \\

w/o representation loss
& 8466 & 1247
& 6241 & 706 \\

w/o adaptive $\beta$ scaling
& 4974 & 830
& 5765 & 259 \\

uniform kernel weighting
& 0 & 0
& 218 & 21 \\


\bottomrule
\end{tabular}
\end{table}






\section{Discussion}
\label{sec:Discussion}

\paragraph{Computation and Complexity.}
Relative to standard off-policy actor--critic methods such as SAC, SAVGO introduces extra per-update cost by sampling and scoring $K$ candidate actions per state in order to form the similarity kernel (Eqs.~\ref{eq:kernel_weights}--\ref{eq:q_tilde}). Let $B$ denote the minibatch size, $C_Q$ the cost of one critic forward pass, and $C_z$ the cost of one encoder forward pass. Ignoring constants and minor bookkeeping, a SAC-style update requires $\mathcal O(B\,C_Q)$ critic evaluation for targets and losses (plus $\mathcal O(B)$ policy log-prob terms). In SAVGO, the critic update remains unchanged up to constants, but the policy improvement step additionally evaluates the critic and encoder on $K$ candidates per state, yielding an overhead of
$\mathcal O\big(BK(C_Q + C_z)\big)$
on top of the baseline critic update. Computing cosine similarities and the softmax weights adds only a few more arithmetic computations, which are typically dominated by network forward passes. Thus, SAVGO's runtime scales approximately linearly in $K$, and the relative overhead is most pronounced when $C_Q$ and $C_z$ are large (e.g., high-dimensional tasks). Memory overhead remains modest, as candidates and similarities are computed on-the-fly and the replay buffer continues to dominate storage.

\paragraph{Assumptions, Practical Considerations, and Limitations.}
SAVGO assumes that the critic provides a reasonably consistent ranking over sampled candidate actions. Under severe critic noise, the resulting similarity weights can become unreliable even with temperature smoothing. The geometry objective likewise assumes that normalized value gaps yield stable supervision, motivating robust scaling ($\beta$) and curvature control ($\lambda$), since overly sharp mappings can amplify noise while overly smooth mappings can obscure action-ranking structure. Performance further depends on candidate coverage. In detail, small $K$ may under-sample the local neighborhood, whereas large $K$ increases computation with diminishing returns, suggesting that $K$ should be tuned relative to action-space dimensionality. Empirical gains are smallest on easy tasks where all methods reach near-ceiling performance, and scaling to very high-dimensional actions may require a larger $K$ with proportional cost. Finally, extending SAVGO to discrete-action domains is non-trivial, as learning a useful state--action geometry and designing effective candidate proposals over discontinuous actions likely requires different encoders and sampling strategies.

\section{Conclusion}
\label{sec:conclusion}

In this paper, SAVGO was introduced as a geometry-aware off-policy actor--critic method in which state--action similarity is made directly actionable during policy improvement. On the MuJoCo continuous-control suite under a fixed 1M-step budget, consistent gains in sample efficiency and stability were observed, with the greatest improvements on challenging high-dimensional tasks such as \textit{Humanoid} and \textit{Walker2d}. Ablation results further verified that value-aware geometry learning, conservative candidate evaluation, adaptive kernel sharpness, and robust value-gap normalization each contribute materially to performance, and removing any component leads to clear degradations or failure cases. Future work will focus on more structured candidate proposal mechanisms for improved coverage in very high-dimensional action spaces, increased robustness under stronger stochasticity or partial observability, and extensions to discrete-action domains (e.g., Atari) through geometry-aware encoders and candidate selection tailored to discontinuous action sets.

\section*{Acknowledgment}
The study was funded by the DriVe2X research and innovation project from the European Commission with grant number 101056934. The authors acknowledge the use of computational resources of the DelftBlue supercomputer, provided by Delft High Performance Computing Centre (https://www.tudelft.nl/dhpc). This work also used the Dutch national e-infrastructure with the support of the SURF Cooperative, using grant no. EINF-5716.


\bibliographystyle{named}
\bibliography{main}

\end{document}